\documentclass{article}

\usepackage{arxiv}

\usepackage[utf8]{inputenc} 
\usepackage[T1]{fontenc}    
\usepackage{hyperref}       
\usepackage{url}            
\usepackage{booktabs}       
\usepackage{amsfonts}       
\usepackage{nicefrac}       
\usepackage{microtype}      
\usepackage{lipsum}		
\usepackage{graphicx}
\usepackage{natbib}
\usepackage{doi}

\title{Bornon: Bengali Image Captioning with Transformer-based Deep learning approach}

\author{{\hspace{1mm}Faisal Muhammad Shah}\\
	Department of Computer Science and Engineering\\
	Ahsanullah University of Science and Technology\\
	Dhaka, Bangladesh \\
	\texttt{faisal.cse@aust.edu} \\
	\And
	{\hspace{1mm}Mayeesha Humaira} \\
	Department of Computer Science and Engineering\\
	Ahsanullah University of Science and Technology\\
	Dhaka, Bangladesh \\
	\texttt{mayeeshahumaira@gmail.com} \\
	
	\And
	{\hspace{1mm}Md Abidur Rahman Khan Jim} \\
	Department of Computer Science and Engineering\\
	Ahsanullah University of Science and Technology\\
	Dhaka, Bangladesh \\
	\texttt{jimrahman33@gmail.com} \\
	
	\And
	{\hspace{1mm}Amit Saha Ami} \\
	Department of Computer Science and Engineering\\
	Ahsanullah University of Science and Technology\\
	Dhaka, Bangladesh \\
	\texttt{amitsaha.aust@gmail.com} \\
	
	\And
	{\hspace{1mm}Shimul Paul} \\
	Department of Computer Science and Engineering\\
	Ahsanullah University of Science and Technology\\
	Dhaka, Bangladesh \\
	\texttt{shimulpaul59@gmail.com} \\
}

\renewcommand{\shorttitle}{\textit{arXiv} Template}

\hypersetup{
pdftitle={A template for the arxiv style},
pdfsubject={q-bio.NC, q-bio.QM},
pdfauthor={David S.~Hippocampus, Elias D.~Striatum},
pdfkeywords={First keyword, Second keyword, More},
}

\begin{document}
\maketitle

\begin{abstract}
	Image captioning using Encoder-Decoder based approach where CNN is used as the Encoder and sequence generator like RNN as Decoder has proven to be very effective. However, this method has a drawback that is sequence needs to be processed in order. To overcome this drawback some researcher has utilized the Transformer model to generate captions from images using English datasets. However, none of them generated captions in Bengali using the transformer model. As a result, we utilized three different Bengali datasets to generate Bengali captions from images using the Transformer model. Additionally, we compared the performance of the transformer-based model with a visual attention-based Encoder-Decoder approach. Finally, we compared the result of the transformer-based model with other models that employed different Bengali image captioning datasets.
\end{abstract}


\section{Introduction}
Image captioning is a process of portraying an image that is done by combining two fields of deep learning. These fields are computer vision and natural language processing (NLP). For many years researchers have examined methods to caption images automatically. This method involves recognizing the objects, attributes, and their relationships with the corresponding images to correctly generate fluent sentences. This is a very challenging task. Image captioning can be used for social and security purposes. It can be used for increasing children’s interest in early education or Security camera footage can be captioned in real-time to prevent theft or prevent any hazard like fire.

The image caption is a sequence modeling problem that employs a CNN-RNN-based encoder-decoder framework. In this task, the encoder is used to extract the image feature to obtain feature vectors, then pass it through an RNN to generate the language description. Previously all researchers utilized this CNN-RNN \cite{25}, \cite{26}, \cite{30} approach to generate captions from images. However, this method has a drawback that is due to the structure of the LSTM or
other RNNs, the current output depends on the hidden state at the previous moment. As a result, they can only operate in time steps, this makes it implausible to parallelize the process of generating the captions. Nevertheless, Vaswani et al. propose the Transformer \cite{1} model solved the parallelism problem. The Transformer can run
in parallel during the training phase as it is based on an attention mechanism there is no sequence dependence on this model. 

\begin{figure}[h!]
	\centering
	\includegraphics[width=0.95\linewidth]{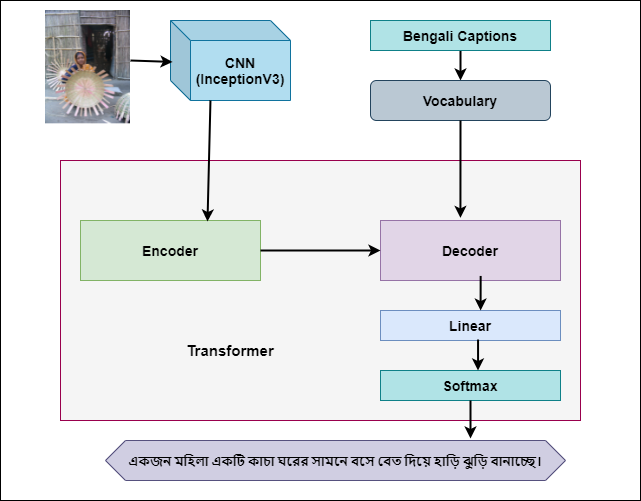}
	\caption{Visualization of how the Transformer model generates words from an input image. First of all, image features extracted were by the CNN and passed to the Encoder of the Transformer. Then the vocabulary was passed to the Decoder part of the Transformer. The Transformer then generated a Bengali caption for the corresponding image.}
	\label{fig7}
\end{figure}

Recently, some researchers \cite{28}, \cite{29} have utilized the Transformer model instead of an RNN to generate captions from images. But, these researches were conducted on English datasets. To see how this model performs in the Bengali dataset we utilized three Bengali datasets. The approach to caption image in Bengali using the transformer model is illustrated in Fig. \ref{fig7}. Furthermore, we compare the performance with the visual attention-based approach to caption images in Bengali that was proposed by Ami et al. \cite{31}. This visual attention-based approach is shown in Fig. \ref{figatt}.  Bengali is the $7^{th}$ most used language worldwide\footnote{\url{ https://www.vistawide.com/languages/top\_30\_languages.htm}} and most of the natives in some parts of India and Bangladesh do not know English. Hence, it is also necessary to caption images in Bengali alongside English. The contributions of this paper are as follows: 

\begin{itemize}
    \item Three Bengali dataset used to train the model.
    \item Transformer model combined with CNN to generate captions from images in Bengali.
    \item Employed a visual attention-based approach to compare its performance with the transformer-based approach.
    \item Compared the performance of other models and the transformer-based model to caption images in Bengali.
\end{itemize}

\begin{figure}[h!]
	\centering
	\includegraphics[width=0.95\linewidth]{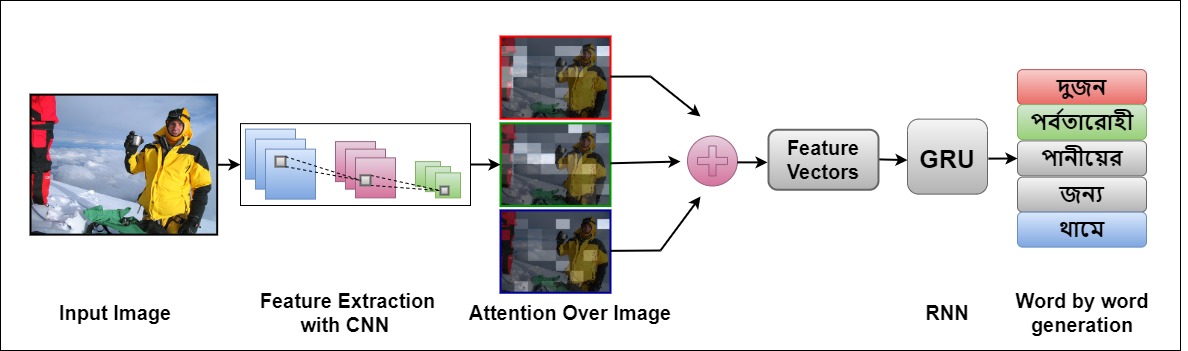}
	\caption{Illustration of how our model learns words from input image to generate caption using visual attention-based approach \cite{31}. First of all, image features were extracted using CNN. Then Attention scores were given to the image features and then passed to the GRU. On the other hand, tokenized words were passed to the embedding layer to covert vocabulary to vectors. These word vectors were also passed to the GRU. The GRU then generates Bengali captions word by word using word vectors and Attention weighted image features.
}
	\label{figatt}
\end{figure}

\section{Related Works} \label{sec2}
This section depicts the progress in image captioning. Hitherto, many types of research have been conducted and many models have been developed in order to get captions that are syntactically corrected.

\subsection{Image captioning in Bengali}
Only seven works have been done on image captioning in Bengali till now. \cite{13} was the first paper in image captioning in Bengali followed by \cite{14}, \cite{15}, \cite{16} and \cite{17}. Rahman et al. \cite{13} have aimed to outline an automatic image captioning system in Bengali called 'Chittron'. Their model was trained to predict Bengali caption from input images one word at a time. The training process was carried out on 15700 images of their own dataset BanglaLekha. In their model Image feature vector and words were converted to vectors after passing them through the embedding, the layer was fed to the stacked LSTM layer. One drawback of their work was that they utilized the sentence BLEU score instead of the Corpus BLEU score. On the other hand, Deb et al. \cite{14} illustrated two models Par-Inject Architecture and Merge Architecture for image captioning in Bengali. In the Par-Inject model image, feature vectors were fed into intermediate LSTM and the output of that LSTM and word vectors were combined and fed to another LSTM to generate caption in Bengali. Whereas, in the Merge model image feature vectors and words vector were combined and passed to an LSTM without the use of an intermediate LSTM. They utilized 4000 images of the Fickr8k dataset and the Bengali caption their models generated were not fluent. Paper \cite{15} used a CNN-RNN based model where VGG-16 was used as CNN and LSTM with 256 channels was used as RNN. They trained their model on the BanglaLekha dataset having 9154 images. On the other hand, paper \cite{16} proposed a CNN-ResNet-50 merged model, consisting of a ResNet-50 as image feature extractor and 1D-CNN with word embedding for generating linguistic information. Later, these two features were given as inputs to a multimodal layer that predicts what to generate next
using the information at each time step. Furthermore, \cite{17} utilized the BNLIT dataset to implement a CNN-RNN model where they used both BRNN and LSTM as RNN. M. Humaira et al. \cite{30} proposed a hybridized Encoder-Decoder approach where two word embeddings fastText and GloVe were concatenated. They also utilized beam search and greedy search to compute the BLEU scores. Additionally, A. S. Ami et al. \cite{31} employed visual attention with the Encoder-Decoder approach to caption images in Bengali. They added attention weights to image features and passed them to the GRU with word vectors to generate captions. However, they did not use corpus BLEU scores to evaluate the captions. We will compare the corpus BLEU scores of the visual attention-based approach with the transformer-based approach.

\subsection{Image captioning in other Languages}
Previously many research was conducted on English as the available datasets were all in the English language. The authors in \cite{23} adapted the attention mechanism to generate caption. For vision part of image captioning VGG-16 were used by most of the papers \cite{24}, \cite{25}, \cite{26} as CNN but some of them also used AlexNet \cite{24}, \cite{26} or ResNet \cite{24} as CNN for feature extraction. However, some of the researchers also utilized BiLSTM \cite{24}. Alongside English researchers also generated captions in Chinese \cite{21}, \cite{22}, Japanese \cite{18}, Arabic \cite{19} and Bahasa Indonesia \cite{20}.

\subsection{Image captioning using Transformer}
The transformer model was used previously for image captioning using an English dataset. Li et al \cite{5} investigated a Transformer-based sequence modeling framework for image captioning which was built only with attention layers and feedforward layers. Additionally, paper \cite{6} employed object spatial relationship modeling for image captioning, specifically within the Transformer encoder-decoder architecture by incorporating the object relation module within the Transformer encoder. Paper \cite{7} proposed the use of augmentation of image captions in a dataset including augmentation using BERT to improve a solution to the image captioning problem. Furthermore, paper \cite{8} utilized two streams of transformer-based architecture. One for the visual part and another for the textual part. Paper \cite{27} used a transformer-based architecture that consists of an encoder and decoder model where the encoder part is a CNN model and the decoder part is a transformer model. It also uses a stacked self-attention mechanism. Paper \cite{28} uses the CNN as an encoder to extract image features, the output of the encoder is a context vector that contains the necessary information from the image, then put it into Transformer to generate the captions. On the other hand, paper \cite{29} introduced the image transformer for image captioning, where each transformer layer implements multiple sub-transformers, to encode spatial relationships between image regions and decode the diverse information in the image regions.

\subsection{Image captioning using Attention Mechanism}
Visual attention on English datasets was used previously by many researchers. In the past two main types of attention were used by researchers in encoder-decoder for image or video captioning. These two types of attention are Semantic attention that is using attention in text and Spatial attention which is applying attention to images. Xu et al. \cite{9} proffered the first visual attention model in image captioning. They used “hard” pooling that designates the most probably attentive region, or “soft” pooling that averages the spatial features with attentive weights. Additionally, Chen et al. \cite{12} utilized Spatial attention and Channel wise Attentions in a CNN. Paper \cite{10} also employed visual attention to generating captions. Lastly, paper \cite{11} employed a semantic attention model to combine the visual feature with visual concepts in a recurrent neural network that generates the image caption.

\section{Model Architecture} \label{sec3}
We utilized the transformer model and the attention-based model proposed by \cite{31} to caption images in Bengali. The transformer model does not process sequence in order but the attention-based model processes sequence in order. Hence, the transformer model allows parallel processing of captions. The transformer model is illustrated in Fig. \ref{fig1} and the attention-based approach is shown in Fig. \ref{figatt}.

\subsection{Transformer-based Approach}
\begin{figure}[h!]
	\centering
	\includegraphics[width=.72\linewidth]{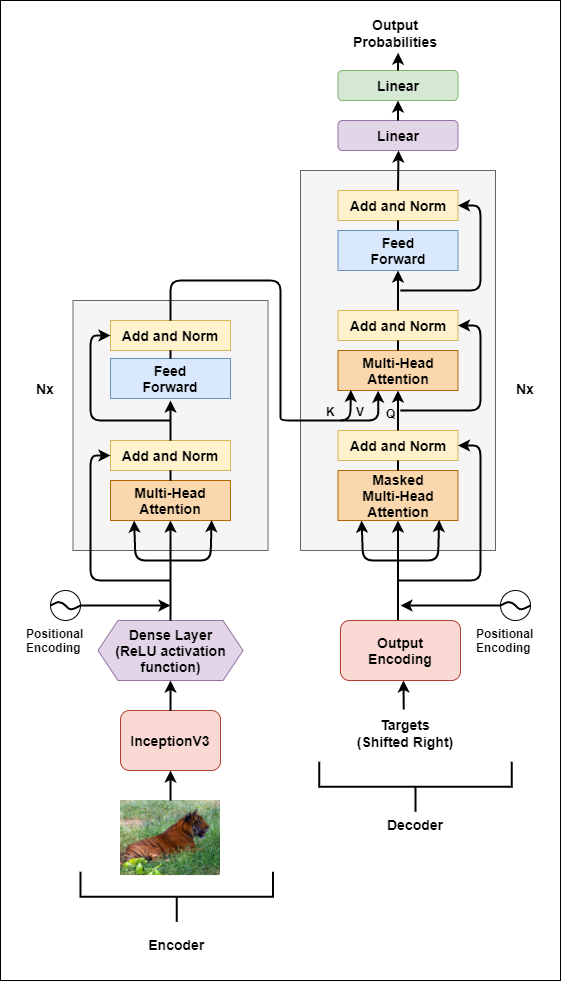}
	\caption{Illustration of the transformer based model to caption image in Bengali.}
	\label{fig1}
\end{figure}

Transformer \cite{1} is a deep learning model that utilizes the mechanism of attention, to give weights to the influences to different parts of the input data. The transformer is made of a stack of encoder and decoder components. In Fig. \ref{fig1} left block marked $N_{x}$ is the encoder and the right block marked $N_{x}$ is the decoder. Here N is a hyperparameter that represents the number of encoder and decoder components.  This model takes two inputs these are image features extracted by the CNN in the Encoder and the vocabulary formed from the list of target captions in the dataset in the Decoder.

\begin{figure}[h!]
	\centering
	\includegraphics[width=.72\linewidth]{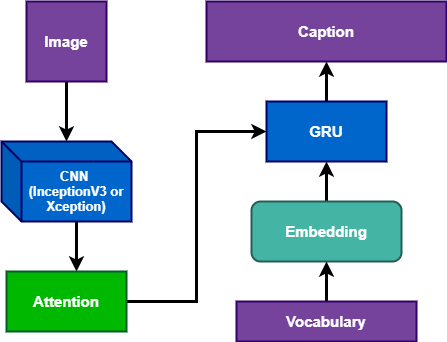}
	\caption{Illustration of the attention-based approach \cite{31} to caption image in Bengali.}
	\label{figatt}
\end{figure}

\subsubsection{Encoder}
InceptionV3 was used as the CNN in this experiment. The images in the dataset were at first passed to the CNN. InceptionV3 extracts image features and passes them through a dense layer having ReLU as an activation function to take the dimension of the image feature vector from d to $d_{model}$ where  $d_{model}$ is the dimension of the word embedding. These image feature vectors were then summed with the positional encoding then passed through N encoder layers. Each of these encoder layers was made up of two sublayers one of which is Multi-head attention with padding mask and the other is Point wise feed-forward networks. Masking ensures that the model does not treat padding as the input. The output of the encoder was then passed to the decoder as K (key) and V (Value). The Multi-Head mechanism is explained in Section \ref{sec7}.

\subsubsection{Decoder}
The decoder takes as input the target captions in the dataset were passed through an embedding which was summed with the positional encoding. Positional encoding is added to give the model some information about the relative position of the words in the sentence based on the similarity of their meaning and their position in the sentence, in the d-dimensional space. The output of the summation was then passed through N decoder layers. Each of these decoder layers was made of three sub-layers one of which was the Masked multi-head attention with a look ahead mask and padding mask, another one was Multi-head attention with padding mask where V (value) and K (key) receive the encoder output as inputs and Q (query) received the output from the masked multi-head attention sublayer. The third layer was Point wise feed-forward networks. The output of the decoder was then sent to the linear layer as input. Finally, using probabilistic softmax predictions one word at a time, and uses the output so far to decide what to do next. This whole process is illustrated in Fig. \ref{fig1}.

\subsection{Visual Attention-based Approach}
To focus only on the relevant parts of the image visual attention-based model was used. It is an Encoder-Decoder approach that processes sequence in order. This model has three main parts. Firstly, a Convolutional Neural Network (CNN) extracts features from images. Secondly, an attention mechanism was utilized to give weights to image features. Bengali vocabulary was then converted to word vectors using an embedding layer. Finally, Gated Recurrent Units (GRU) \cite{35} which is a sequence generator took word vectors and weighted images features as input and generated Bengali captions in order. This process is illustrated in \ref{figatt}.

\section{Dataset} \label{sec4}
The main aim of this research is to generate a Bengali caption from the image. To accomplish this task a dataset in the Bengali language is needed which must have several images and a text file containing Bengali captions associated with each image. However, all the datasets available for image captioning are in English. The only available Bengali dataset till now is the BanglaLekha dataset. Since one dataset is not enough to validate the performance of the models we created a new Bengali dataset named Bornon. Furthermore, we utilized the Flickr8k dataset by translating its English captions to Bengali and then merging it with the BanglaLekha and Bornon dataset to form a newly merged dataset. This merged dataset was constructed especially to test the transformer model since the transformer-based models are data-hungry. For generating Bengali cations from images, these datasets were split into three parts: training, testing, and validation. The split ratio of each dataset used in our experiment is shown in Table \ref{tab1}.

\begin{table}
\caption{Distribution of Data for Different Bengali Dataset used in our experiment.}
\label{table}
\setlength{\tabcolsep}{19pt}
\begin{tabular}{c c c c c }
\hline
\hline
\textbf{Dataset} 
&\textbf{Total Image} 
&\textbf{Training}
&\textbf{Validation} 
&\textbf{Testing}\\
\hline
 Flickr8k & 8000 & 6000 (75\%) & 1000 (15\%) & 1000 (15\%) \\ \hline 
BanglaLekha & 9154 & 7154 (78\%) & 1000 (11\%) & 1000 (11\%) \\\hline
Bornon & 4100 & 2900 (72\%) & 600 (14\%) & 600 (14\%) \\ \hline 
merged & 21414 & 12850 (60\%) & 4282 (11\%) & 4282 (11\%) \\\hline
\end{tabular}
\label{tab1}
\end{table}

\subsection{Flickr8K\_BN}
\begin{figure}[h!]
	\centering
	\includegraphics[width=1.0\linewidth]{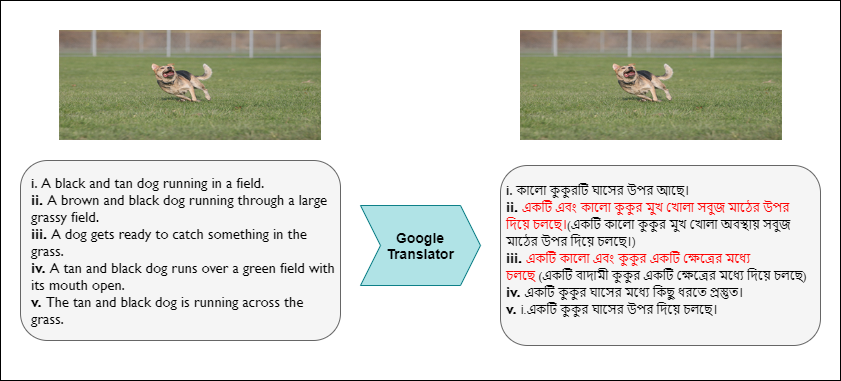}
	\caption{Illustration of Bengali captions after being translated using Using Google Translator. Sentences marked with red color indicate syntactically incorrect Bengali sentences and the sentences inside the brackets are the manually corrected sentences.}
	\label{figflickr}
\end{figure}

\begin{figure}[h!]
	\centering
	\includegraphics[width=1.0\linewidth]{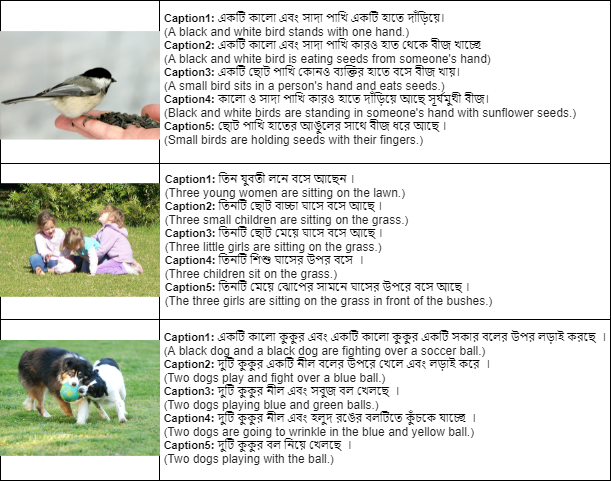}
	\caption{Illustration of some images of the Flickr8k dataset along with their five Bengali captions.}
	\label{fig2}
\end{figure}

Flickr8k\footnote{https://www.kaggle.com/adityajn105/flickr8k/activity} dataset is a publically available English dataset that contains 8091 images of which 6000 (75\%) images are employed for training, 1000 (12.5\%) images for validation, and 1000 (12.5\%) images are used for testing. Moreover, with each image of the Flickr8K dataset five ground truth captions describing the image are designated which adds up to a total of 40455 captions for 8091 images. For image captioning in Bengali, those 40455 captions were converted to Bengali language using Google Translator\footnote{https://translate.google.com/}. Unfortunately, some of the translated captions were syntactically incorrect as shown in Fig \ref{figflickr}. Hence, we manually checked all 40455 translated captions and corrected them. Flickr8K-BN is the Bengali Flickr8K dataset. Some images of the Flickr8k dataset along with their associated captions are shown in Fig. \ref{fig2}.

\subsection{BanglaLekha}
\begin{figure}[h!]
	\centering
	\includegraphics[width=1.0\linewidth]{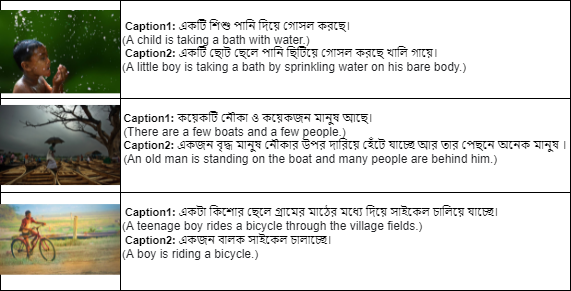}
	\caption{Illustration of some images of the BanglaLekha dataset along with their two Bengali captions.}
	\label{fig3}
\end{figure}

We also utilized the BanglaLekha\footnote{https://data.mendeley.com/datasets/hf6sf8zrkc/2} dataset which consists of 9154 images of which 7154 (78\%) images are employed for training, 1000 (11\%) images for validation, and 1000 (11\%) images are used for testing. It is the only available Bengali dataset till now. All its captions are human-annotated. One problem with this dataset is that it has only two captions associated with each image resulting in 18308 captions for those 9154 images. Hence, vocabulary size is lower than Flickr8k-BN. Flickr8k-BN consists of 12953 unique Bengali words, and BanglaLekha consists of 5270 unique Bengali words. It can be seen that the BanglaLekha dataset has a vocabulary size even lower than Flickr8k-BN. Some images of the BanglaLekha dataset along with their associated captions are shown in Fig. \ref{fig3}.

\subsection{Bornon}
\begin{figure}[h!]
	\centering
	\includegraphics[width=1.0\linewidth]{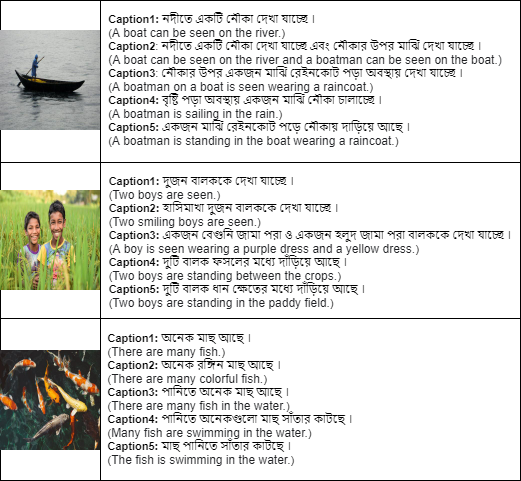}
	\caption{Illustration of some images of the Bornon dataset along with their five Bengali captions.}
	\label{fig4}
\end{figure}

Due to the lack of a Bengali image captioning dataset and to overcome the shortcomings of the existing Bengali image captioning datasets Banglalekha \cite{13}, we created a new dataset named Bornon. The Bornon dataset consists of 4100 images and each image has five captions describe them. Thus, there is a total of 20500 captions for 4100 images. Images were kept in a folder and the associated captions were kept in a text file. Some images of the Bornon dataset along with their associated captions are shown in Fig. \ref{fig4}.

\begin{figure}[h!]
	\centering
	\includegraphics[width=1.0\linewidth]{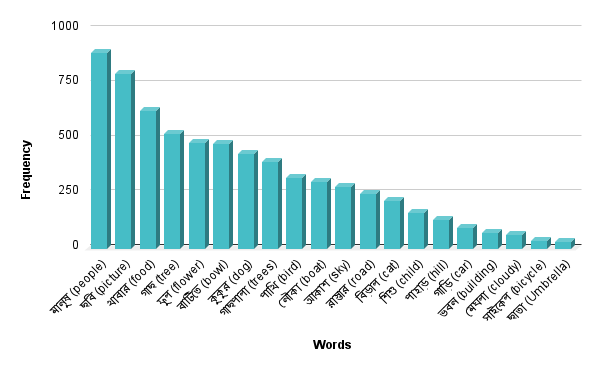}
	\caption{Illustration most frequent Bengali words in the Bornon Dataset.}
	\label{figchart}
\end{figure}

The images of this dataset were taken from a personal photography club. All images were in jpg format. These images portray various objects like Animals, Birds, People, Food, Weather, Trees, Flower, Buildings, Cars, Boat. Frequent Bengali words in this dataset are illustrated in Fig. \ref{figchart}. Around 17 people who are native Bengali speakers were responsible for annotating and evaluating the captions.

 Only two captions are associated with each image in the BanglaLekha dataset this reduced the vocabulary size hence we gave five captions for each image in our Bornon dataset. The vocabulary size of the Bornon dataset was 6228 unique Bengali words for only 4100 images whereas the BanglaLekha dataset had a vocabulary size of  5270 unique Bengali words for 9154 images. If vocabulary size is the small repetition of words is observed in predicted captions. However, this 4100 data is not enough to train a transformer-based mode. Therefore, in the future, we plan to increase the number of images in our dataset.

\subsection{Merged Dataset}
The transformer model is data-hungry. It performs well when a huge number of data is provided. However, all the Bengali datasets mentioned above have a small amount of data which is not enough to train the transformer model. As a result, we merged three datasets Flickr8k, BanglaLekha, and Bornon. This resulted in 21414 images and each image had two captions associated with them which add up to a total of 42828 captions. We took two captions from all the datasets because the BanglaLekha dataset had only two Bengali captions describing each image. This merging led to a vocabulary size of 13416 unique Bengali words and images of various categories.

\section{Text Embedding} \label{sec5}
Firstly, the maximum length of target captions in each dataset was computed. Then all the sentences having lengths less than maximum length were padded with zeroes. Afterward, the top 5000 unique Bengali words were selected from each dataset to tokenize the Bengali captions using Keras’s text tokenizer. Since we cannot train a model using text we converted these tokens to numeric form using text embedding. The embedding model embeds these tokens to one-hot vectors having d\_modeldimensions. All these embedding vectors in one sentence are combined into a matrix and provided as input to the Transformer or the GRU.

\section{Convolutional Neural Network} \label{sec6}
InceptionV3 \cite{2} was utilized as the Convolutional Neural Network (CNN) for extracting features from images for the transformer-based model. As this is not a classification task, the last layer of InceptionV3, a softmax layer, was removed from the model. Then all the images were preprocessed to the same size, that is, 299×299 before feeding them into the model. Hence, the shape of the output of this layer was 8x8x2048. The features were extracted and stored as .npy files and then pass those features through the encoder.

Two different CNN InceptionV3 and Xception \cite{32} were employed in the different experimental setups of the visual attention-based model. The last layer of both CNN was removed. Then like the transformer model the attention-based model also took images of size 299x299. As a result, images were reshaped and feed to CNN. The extracted image features were then to .npy files and attention weight was added to them.

\section{Attention in Transformer} \label{sec7}
Self-attention is calculated using vectors. Three matrices Query, Key, and Value from each of the encoder’s inputs are needed to calculate self-attention. These matrices are obtained by multiplying the embedding matrix and the Weight trained weight matrices. Finally, self-attention matrices are calculated using the following formula.

\begin{equation}
Z = softmax(\frac{Q*K^{T}}{\sqrt{d_{k}}}) * V\label{eq1}
\end{equation}

Where Z is the self-attention matrix, Q is the Query matrix, K is the Key matrix, V is the Value matrix $d_{k}$ is the dimension of the key matrix. The paper \cite{1} further refined the self-attention layer by adding a mechanism called “Multi-Head” attention. Multi-Head attention implements the self-attention calculation eight different times with different weight matrices.

\section{Visual Attention Mechanism}
Two types of spatial attention that are used widely are Global attention \cite{34} and Local attention. We employed Local attention which is also known as Bahdanau attention \cite{33} because Global attention is computationally expensive and unfeasible for large sentences. Global attention place attention on all source position whereas Bahdanau attention focuses on a small subset of hidden states of the encoder per target word. To implement the Bahdanau Attention several steps have been followed. Firstly, the extracted image features were passed through a Fully connected layer using CNN Encoder to produce a hidden state of each element. Then Alignment scores were calculated using the hidden state produced by the decoder in the previous time step and the encoder outputs using the formula shown in Eq. \ref{eq25}. This Alignment score is the main component of the attention mechanism.

\begin{equation}
        score_{alignment} = W_{combined}.tanh(W_{decoder}.H_{decoder}+W_{encoder}.H_{encoder})\label{eq25}
    \end{equation}
    
    The Alignment Scores were then passed through the SoftMax function and represented in a single vector called attention weights using Eq. \ref{eq26}. This vector was then multiplied with image features to form the context vector using Eq. \ref{eq27}.
    
    \begin{equation}
        a_{jt} = Softmax(e_{jt})
    \label{eq26}
    \end{equation}

Where, $a_{jt}=\frac{e^{e_{jt}}}{\sum_{k=1}^{T_{x}}e^{e^{kt}}}$  ,such that  $\sum_{j = 1}^{T_{x}} a_{jt} = 1$ and $a_{ij}\geq 0 $ and $e_{jt}$ is the score-alignment.

    \begin{equation}
        c_{t}=\sum _{j = 1}^{T_{x}} \alpha_{jt}h_{j}\label{eq27}
    \end{equation}

Where $a_{jt}=\sum_{j = 1}^{T_{x}} a_{jt} = 1 $ and $a_{ij}\geq 0 $ and $c_{t}$ is the context vector that is the weighted sum of the input.\newline

Finally, this context vector was concatenated with the previous decoder output. It was then fed into the decoder Gated Recurrent Unit (GRU) to produce a new output.

\section{Gated Recurrent Units}
In the attention-based approach, Gated Recurrent Units  (GRU)\cite{35} was employed as a sequence generator.  Before passing words to the GRU they were converted to vectors using the embedding layer. Afterward, this word embedding of Bengali words was passed to GRU. The GRU then predicts the next word in the sequence using the previous hidden state of the decoder, the previous predicted word, and the context vector calculated in the attention model. The equation used to predict the next word is depicted in Eq. \ref{eq28}. 
\begin{equation}
    s_{t} = RNN(s_{t-1},[e(\hat{y}_{t-1}),c_{t}])\label{eq28}
\end{equation}
Where,$s_{t}$ is the new state of the decoder, $s_{t-1}$is previous state of decoder, e($\hat{y}_{t-1}$)is previous predicted word and
$c_{t}$ is the context vector that is the weighted sum of the input.

However, the sequence problem remains. We need to process the data so that is we need to process the beginning of the sequence before the end. To solve this issue we utilized the transformer-based model to caption images in Bengali.

\section{Hyperparameters} \label{sec8}
The techniques used in this experiment were implemented by Jupyter Notebook. These models were developed based on Keras 2.3.1 and Tensorflow 2.1.0. We ran our experiments on NVIDIA RTX 2060 GPU. RTX 2060 offers 1920 CUDA cores with 6 GB GDDR6 VRAM. Using these settings it took approximately three hours to train each of the experimental setups. 

\begin{figure}[h!]
	\centering
	\includegraphics[width=0.5\linewidth]{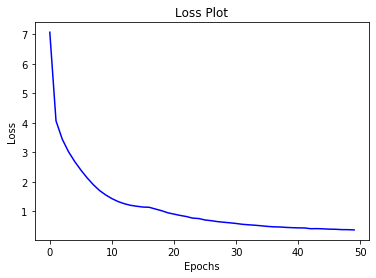}
	\caption{Loss plot for transformer-based model over 50 epoch using 3 layers and 2 heads for Bornon dataset.}
	\label{fig5}
\end{figure}

The number of layers and the number of heads in the transformer were varied to tune the transformer-based model. Three, five, and seven were used as some layers in the transformer and one and two were used as the number of heads of the transformer.  Furthermore, Internal validation was employed in this model to test the generalization ability of our trained model. Moreover, this model was trained for 50 epochs and utilized the Adam optimizer with a custom learning rate scheduler according to the Eq. \ref{eq2} where 4000 was used as the warmup\_steps. Additionally, SparseCategoricalCrossentropy was utilized as the loss function. The loss plot of one of the experimental setups using the transformer-based model is shown in Fig. \ref{fig5}.

\begin{equation}
l_{rate} = d^{-0.5}_{model} * min(step\_num^{-0.5},\quad step\_num * warmup\_steps^{-1.5}  )\label{eq2}
\end{equation}

\begin{figure}[h!]
	\centering
	\includegraphics[width=0.5\linewidth]{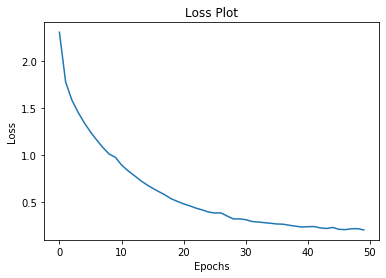}
	\caption{Loss plot for visual attention-based model using InceptionV3 over 50 epoch using Flickr8k-BN dataset.}
	\label{figincloss}
\end{figure}

\begin{figure}[h!]
	\centering
	\includegraphics[width=0.5\linewidth]{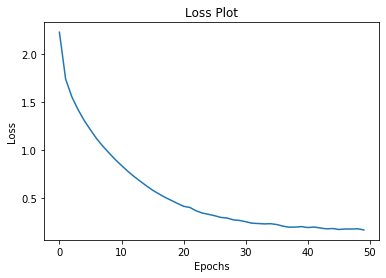}
	\caption{Loss plot for visual attention-based model Xception over 50 epoch using Flickr8k-BN dataset.}
	\label{figxcloss}
\end{figure}

Two different CNN InceptionV3 and Xception were used in the different experimental setups as hyperparameters in the visual attention-based model. Furthermore, Internal validation was employed in this model to test the generalization ability of our trained model. Moreover, this model was trained for 50 epochs and had a batch size of 64. Additionally, Adam optimizer was used as the optimizer and for calculating the loss SparseCategoricalCrossentropy was utilized. Fig. \ref{figincloss} and Fig. \ref{figxcloss} demonstrates how loss decreased over 50 epoch for InceptionV3 and Xception respectively.

\section{Experimental Results} \label{sec9}
After generating captions, the most important part is evaluating them to verify how similar the generated captions are to human-annotated captions. Hence, we took the aid of two evaluation metrics BLEU and METEOR to justify the accuracy of our proposed models. 

\subsection{BLEU}
Bilingual Evaluation Understudy (BLEU) \cite{4} is the most wielded metric nowadays to evaluate the merit of text. It depicts how normal sentences are compared with human-generated sentences.  It is predominantly utilized to evaluate the performance of Machine translation. Sentences are compared based on modified n-gram precision for generating BLEU scores. BLEU scores are computed using the following equations:
\begin{equation}
P(i) = \frac{Matched(i)}{H(i)}\label{eq3}
\end{equation}

P(i) is the precision for each i-gram where i = 1, 2, ...N, the percentage of the i-gram tuples in the hypothesis that also occurs in the references is computed. H(i) is the number of i-gram tuples in the hypothesis and Matched(i) is computed using the following formula:

\begin{equation}
Matched(i) = \sum_{t_{i}}\min{\{C_{h}(t_{i}), \max_{j}C_{hj}(t_{i})\}}\label{eq4}
\end{equation}

where $t_{i}$ is an i-gram tuple in hypothesis h, $C_{h}(t_{i})$ is the number of times $t_{i}$ occurs in the hypothesis, $C_{hj}(t_{i})$ is the number of times $t_{i}$ occurs in reference j of this hypothesis. 

\begin{equation}
\rho = exp\{\min(0, \frac{n-L}{n})\}\label{eq5}
\end{equation}

where $\rho$ is the brevity penalty to penalize short translation, n is the length of the hypothesis and L is the length of the reference. Finally, the BLEU score is computed by:

\begin{equation}
BLEU = \rho \{\prod_{i=1}^{N}P(i)\}^{\frac{1}{N}}\label{eq6}
\end{equation}


\subsection{METEOR}
Metric for Evaluation of Translation with Explicit Ordering (METEOR) \cite{3} is based on unigram matching between reference and predicted sentences by machine using the harmonic mean of unigram precision and recall. The recall is weighted higher than precision here. It was formulated to mend some of the issues found in BLEU metrics.  Unigram precision P is calculated as:

\begin{equation}
P = \frac{m}{w_{t}}\label{eq7}
\end{equation}

Where m is the number of unigrams in the candidate translation that are also found in the reference translation, and $w_{t}$ is the number of unigrams in the candidate translation. Unigram recall R is computed as follows:

\begin{equation}
R = \frac{m}{w_{r}}\label{eq8}
\end{equation}

Where m is as above, and $w_{r}$ is the number of unigrams in the reference translation. Precision and recall are combined using the harmonic mean. There recall is weighted 9 times more than precision as shown in the equation below:

\begin{equation}
F_{mean} = \frac{10PR}{R + 9P}\label{eq9}
\end{equation}

To account for congruity concerning larger segments that appear in both the reference and the candidate sentence a penalty p is added. The penalty is calculated using the following equation.

\begin{equation}
p = 0.5 (\frac{C}{u_{m}})^{3}\label{eq10}
\end{equation}

Where C is the number of chunks, and $u_{m}$ is the number of unigrams that have been mapped. Finally, the METEOR score for a segment is calculated as M as shown in the equation below.

\begin{equation}
M = F_{mean} (1-p)\label{eq11}
\end{equation}


\subsection{Result Analysis}
We employed BLEU and METEOR scores for every experimental setup. These scores of the transformer-based model are shown in Table \ref{tab2} and the scores for the visual attention-based model are shown in Table \ref{tab3}. The highest BLEU for all datasets score using the Transformer-based model was obtained using 3 layers. On the other hand, METEOR was higher for the BanglaLekha dataset with 7 layers and higher for the Bornon dataset when 3 layers were used. For the merged dataset METEOR scores did not show any trend of increasing or decreasing with the number of layers. These scores were even better than BLEU scores obtained by paper \cite{14}, paper \cite{15} and paper \cite{16}. Bornon and BanglaLekha datasets performed slightly better than the Merged dataset using the transformer-based method.

From Table \ref{tab3} it can be seen that experimental setups of the visual attention-based model with Xception as CNN gave higher scores. However, the overall BLEU scores are lower than the transformer-based model. As the visual attention-based model used a GRU as a sequence generator whereas the transformer model was used as a sequence generator in the transformer-based model. This proves the fact that only improving the computer vision side of the image captioning models won’t improve the results. Since image captioning is a mixture of two fields computer vision and NLP equal importance must be given to both fields to get better results. In Table \ref{tab3} we illustrated the corpus BLEU scores which were not done by \cite{31}.

\begin{table}
\caption{Result of Transformer-based model using InceptionV3 as CNN over 50 epochs.}
\label{table}
\setlength{\tabcolsep}{12pt}
\begin{tabular}{c c c c c c c c }
\hline
\hline
\textbf{Dataset}& 
\textbf{Layers(N)}& 
\textbf{Heads}&
\multicolumn{4}{c}{\textbf{BLEU}}&
\textbf{METEOR}\\\cline{4-7}
&
&
&
\textbf{1}&
\textbf{2}&
\textbf{3}&
\textbf{4}&
\\
\hline
 &  & 1 & \textbf{0.665} & \textbf{0.556} & \textbf{0.476} & \textbf{0.408} & 0.255 \\ \cline{3-8}
  & 3 & 2 & 0.662 & 0.548 & 0.462 & 0.389 & 0.241 \\ \cline{2-8}
  &  & 1 & 0.648 & 0.546 & 0.470 & 0.402 & 0.251 \\ \cline{3-8}
BanglaLekha  & 5 & 2 & 0.660 & 0.557 & 0.480 & 0.415 & 0.263 \\ \cline{2-8}
  &  & 1 & 0.633 & 0.541 & 0.471 & 0.409 & 0.267 \\ \cline{3-8}
  & 7 & 2 & 0.644 & 0.548 & 0.476 & 0.412 & \textbf{0.268} \\ \hline
  
    &  & 1 & \textbf{0.696} & \textbf{0.589} & \textbf{0.507} & \textbf{0.439} & \textbf{0.361} \\ \cline{3-8}
  & 3 & 2 & 0.687 & 0.572 & 0.486 & 0.415 & 0.346 \\ \cline{2-8}
  &  & 1 & 0.688 & 0.583 & 0.502 & 0.437 & 0.359 \\ \cline{3-8}
Bornon  & 5 & 2 & 0.683 & 0.575 & 0.493 & 0.425 & 0.340 \\ \cline{2-8}
  &  & 1 & 0.684 & 0.567 & 0.478 & 0.405 & 0.340 \\ \cline{3-8}
  & 7 & 2 & 0.665 & 0.556 & 0.477 & 0.411 & 0.346 \\ \hline
  
    &  & 1 & \textbf{0.621} & \textbf{0.492} & \textbf{0.398} & \textbf{0.326} & 0.196 \\ \cline{3-8}
  & 3 & 2 & 0.624 & 0.494 & 0.400 & 0.329 & 0.201 \\ \cline{2-8}
  &  & 1 & 0.616 & 0.482 & 0.384 & 0.311 & 0.189 \\ \cline{3-8}
merged  & 5 & 2 & 0.607 & 0.483 & 0.391 & 0.323 & \textbf{0.200} \\ \cline{2-8}
  &  & 1 & 0.592 & 0.468 & 0.376 & 0.308 & 0.187 \\ \cline{3-8}
  & 7 & 2 & 0.602 & 0.481 & 0.390 & 0.322 & 0.197 \\ \hline
\end{tabular}
\label{tab2}
\end{table}


\begin{table}
\caption{ Result of \textbf{Visual attention-based model} using GRU as sequence generator over 50 epochs.}
\label{table}
\setlength{\tabcolsep}{15pt}
\begin{tabular}{c c c c c c c c }
\hline
\hline
\textbf{Dataset} 
&\textbf{CNN} 
 &\multicolumn{4}{c}{\textbf{BLEU}}
 &\textbf{METEOR}\\\cline{3-6}
\textbf{} 
&\textbf{}
&\textbf{1} 
&\textbf{2} 
&\textbf{3} 
&\textbf{4} 
&\textbf{}\\ 
\hline
Flickr8k-BN   &  InceptionV3 & 0.543 & 0.445& 0.362 & 0.294 & \textbf{0.161} \\ \cline{2-7}
  & Xception & \textbf{0.546} & \textbf{0.447} & \textbf{0.364} & \textbf{0.29}6 & 0.156 \\ \hline
BanglaLekha   &  InceptionV3 & 0.567 & 0.460 & 0.385 & 0.319 & 0.204 \\ \cline{2-7}
  & Xception & \textbf{0.570} & \textbf{0.462} & \textbf{0.387} & \textbf{0.322} & \textbf{0.208} \\ \hline
 Bornon   &  InceptionV3 & 0.596 & 0.475 & 0.390 & 0.324 & 0.314 \\ \cline{2-7}
  & Xception & \textbf{0.605} & \textbf{0.492} & \textbf{0.412} & \textbf{0.351} & \textbf{0.348} \\ \hline
  \end{tabular}
\label{tab3}
\end{table}

\begin{figure}[h!]
	\centering
	\includegraphics[width=1.0\linewidth]{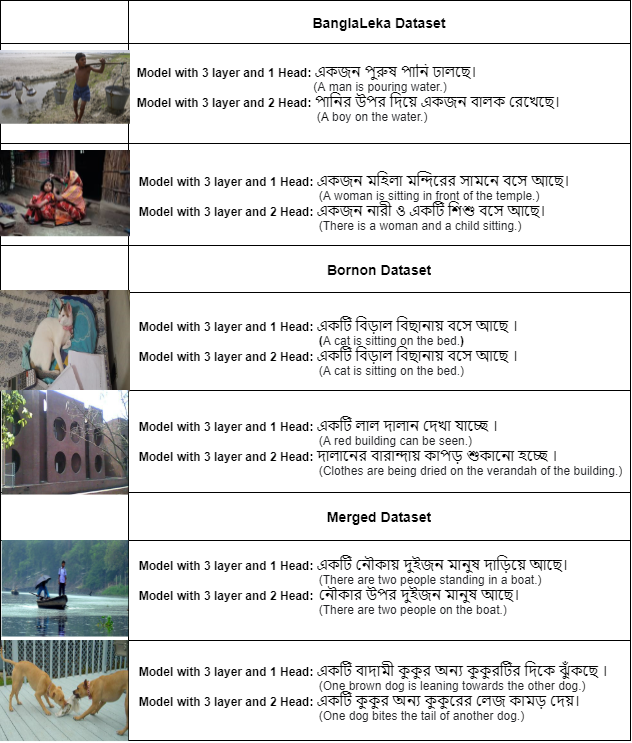}
	\caption{Illustration of Bengali captions generated by Transformer-based models.}
	\label{fig13}
\end{figure}

\begin{figure}[h!]
	\centering
	\includegraphics[width=1.0\linewidth]{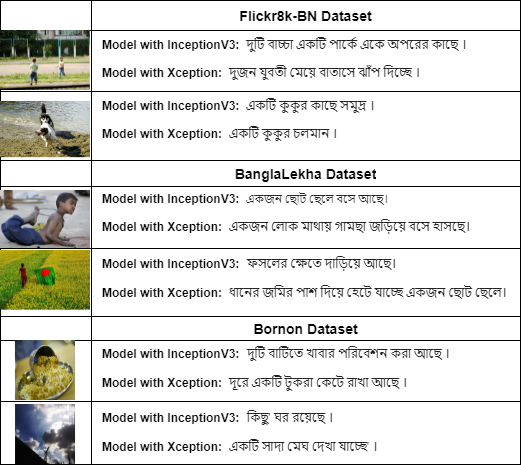}
	\caption{Illustration of Bengali captions generated by visual attention-based models.}
	\label{fig14}
\end{figure}

We tested the transformer-based model and the visual attention-based model using a test set that contains different images that were not present in the training set or validation set. The Bengali captions generated by various experimental setups of the transformer-based model using three datasets BanglaLekha, Bornon, and merged dataset are shown in Fig. \ref{fig13}. Additionally, the Bengali captions generated by the visual attention-based model using Flickr8k-BN, BanglaLekha, and Bornon datasets are illustrated in Fig. \ref{fig14}.  From these figures, it can be seen that the transformer-based model gave much better and accurate Bengali captions than the attention-based model.

\begin{figure}[h!]
	\centering
	\includegraphics[width=1.0\linewidth]{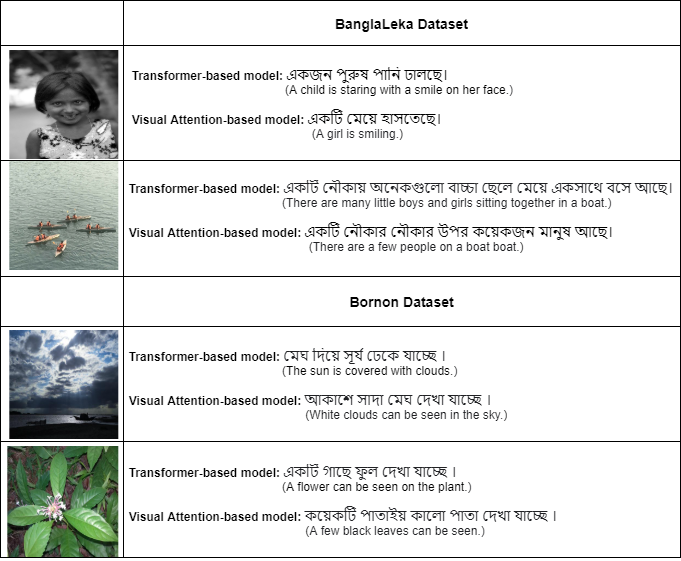}
	\caption{Illustration of Bengali captions generated by visual attention-based models and transformer-based models. Visual attention-based Bengali captions were generated using Xception and the transformer-based Bengali captions were generated using 3 layers and 4 heads.}
	\label{fig15}
\end{figure}

Since both transformer-based model and visual attention-based model were trained using BanglaLekha dataset and Bornon dataset a brief comparison of caption generated for the same test images of these datasets is depicted in Fig. \ref{fig15}. From this figure, it can be seen that the visual attention-based model generated Bengali captions related to the objects present in the caption whereas the transformer-based model gave a general Bengali caption that describes the whole image. Performances of three of the transformer-based model were compared with the performance of other papers and the results are illustrated in Table \ref{tab4}. This table shows that the transformer-based model performed better than other research done on Bengali image captioning using the same datasets.

\begin{table}
\caption{A brief comparison of BLEU scores for existing models and the transformer-based model.}
\label{table}
\setlength{\tabcolsep}{15pt}
\begin{tabular}{c c c c c c c c}
\hline
\hline
\textbf{Dataset}& 
\textbf{Model}& 
\multicolumn{4}{c}{\textbf{BLEU}}\\ \cline{3-6}
&
&
\textbf{1}&
\textbf{2}&
\textbf{3}&
\textbf{4} \\
\hline
& VGG-16+LSTM \cite{15} &  0.667 &0.436 &0.315 &0.238 \\ \cline{2-6}
   BanglaLekha& CNN-ResNet-50 \cite{16} &  0.651 &0.426 &0.278 &0.175 \\ \cline{2-6}
 & Transformer Model & \textbf{0.665} & \textbf{0.556} & \textbf{0.476} & \textbf{0.408} \\\hline\\
 Flickr8k(4000 images) & Inception+LSTM \cite{14}&  0.62 &0.45 &0.33 &0.22\\
 \hline
\end{tabular}
\label{tab4}
\end{table}

\section{Conclusions} \label{sec10}
In our work, we employed a visual attention-based approach that gives attention weight to image features. This was a traditional Encoder-Decoder approach so we compared it with a transformer-based approach. In the transformer-based approach, we combine the feature vector extracted by CNN and target Bengali captions into the Transformer model. This model learns to generate Bengali captions using a multi-head attention mechanism. Not only the model can improve the original performance, but also uplift the training speed by allowing parallelism. Later it was validated that the transformer-based method indeed performs better than the visual attention-based method. Hence, in the future, the transformer-based model can replace the traditional encoder-decoder architecture. This will enhance the performance and efficiency of caption generation from images. We also utilized various Bengali datasets to test both approaches. This proves the fact that the transformer model can be used to generate captions from images in other languages alongside English.

\bibliographystyle{unsrtnat}
\bibliography{references}  






\end{document}